\documentclass[letterpaper, 10 pt, conference]{ieeeconf}  
\IEEEoverridecommandlockouts                            
\usepackage{graphics}
\usepackage{epsfig}
\usepackage{mathptmx}
\usepackage{times}
\usepackage{amsmath}
\usepackage{amssymb}
\usepackage{multirow}
\usepackage{makecell}
\usepackage{newtxtext}
\usepackage{newtxmath}
\usepackage{xurl}
\usepackage{hyperref}
\usepackage{graphicx}    
\usepackage{subcaption}  
\usepackage{algorithmicx}
\usepackage[ruled,vlined, linesnumbered]{algorithm2e}
\usepackage{booktabs}
\usepackage{multirow}

\makeatletter
\newcommand{\removelatexerror}{\let\@latex@error\@gobble}

\title{\LARGE \bf Deep Reinforcement Learning-Enhanced Event-Triggered Data-Driven Predictive Control for a 3D Cable-Driven Soft Robotic Arm}

\author{Cheng Ouyang$^{1}$, Moeen Ul Islam$^{1}$, Kaixiang Zhang$^{2}$, Zhaojian Li$^{2}$, Xiaobo Tan$^{3}$, and Dong Chen$^{1}$
\thanks{$^{1}$Cheng Ouyang, Moeen Ul Islam, and Dong Chen are with the Department of Agricultural \& Biological Engineering, Mississippi State University, MS, USA. {\tt\small Emails: \{co603, mu136, dc2528\}@msstate.edu}}%
\thanks{$^{2}$Kaixiang Zhang and Zhaojian Li are with the Department of Mechanical Engineering, Michigan State University, MI, USA.
        {\tt\small Emails: \{zhangk64, lizhaoj1\}@msu.edu}}%
\thanks{$^{3}$Xiaobo Tan is with the Department of Electrical and Computer Engineering, Michigan State University, MI, USA.
{\tt\small Email: xbtan@egr.msu.edu}}}

\begin{document}

\maketitle
\thispagestyle{empty}
\pagestyle{empty}

\begin{abstract}
Soft robots are challenging to control due to their nonlinear and time-varying dynamics. Data-enabled predictive control (DeePC) offers a model-free alternative by directly leveraging measured input–output trajectories to construct a predictive controller. However, its receding-horizon formulation requires solving a constrained optimization problem at every sampling instant, which can be computationally demanding for real-time deployment on resource-limited robotic platforms. 
To address this limitation, we propose an adaptive reinforcement-learning-based event-triggered DeePC (RL-ET-DeePC) framework for soft robotic control. A model-free RL policy is trained to determine when to invoke the DeePC optimizer based on the current system state representation, thereby reducing unnecessary optimization calls while preserving closed-loop performance. 
Simulation results show that RL-ET-DeePC reduces optimization frequency by up to 66\% compared to periodic DeePC, while maintaining comparable tracking accuracy. Hardware experiments on a three-dimensional cable-driven soft robotic arm demonstrate zero-shot transfer, achieving a 34\% reduction in optimization frequency with tracking accuracy comparable to periodic DeePC and more consistent performance than a static threshold-based event-triggered baseline.
\end{abstract}

\section{INTRODUCTION}
Soft robots have attracted significant attention due to their inherent compliance and safety, enabling applications in minimally invasive surgery \cite{runciman2019soft}, deep-sea exploration of fragile biological specimens \cite{galloway2016soft}, and safe human-robot interaction in shared environments \cite{polygerinos2017soft}. Their ability to undergo continuous deformation allows them to adapt effectively to unstructured and contact-rich tasks \cite{rus2015design}. Despite their versatility, controlling soft robotic arms remains a significant challenge due to the difficulty of developing accurate yet computationally tractable models \cite{thieffry2018control}. Unlike rigid robots, which can be described using standard rigid-body dynamics, soft robots exhibit theoretically infinite degrees of freedom and complex behaviors such as strong nonlinearity and time-varying deformation \cite{trivedi2008soft}.

\begin{figure}
  \centering
  \includegraphics[width=0.80\columnwidth]{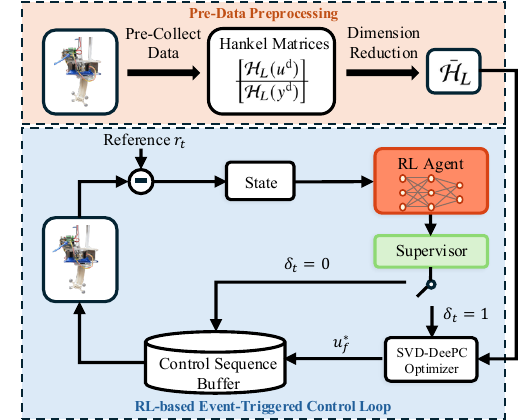}
\caption{Illustration of the proposed RL-ET-DeePC framework. The upper block shows pre-data collection and SVD-based Hankel matrix reduction. 
The lower block depicts the real-time event-triggered control loop, where the RL policy determines whether to trigger the SVD-DeePC optimizer or reuse buffered inputs under supervisory constraints.}
\vspace{-14pt}
\label{fig:framework}
\end{figure}

Traditional approaches, such as Finite Element Methods (FEM) or analytical formulations based on continuum mechanics (e.g., Cosserat rods), can capture soft-robot deformations with high fidelity \cite{till2019real,qin2024modeling,boyer2020dynamics}. However, these methods typically require solving high-dimensional partial differential equations (PDEs), making them computationally prohibitive for real-time control, particularly on embedded platforms \cite{della2023model}. 
To overcome the limitations of explicit physical modeling, data-driven control strategies have emerged as an attractive alternative. Among them, data-enabled predictive control (DeePC)~\cite{coulson2019data} has gained significant attention. By directly leveraging measured input-output trajectories to characterize system behavior, DeePC eliminates the need for parametric model identification while retaining the constraint-handling capabilities of optimization-based control methods, such as model predictive control (MPC) \cite{markovsky2021behavioral}.

Recent studies have explored the application of DeePC to soft robotic control \cite{wang2024mechanical, wang2025velocity, ouyang2025direct}. For instance, Wang et al. \cite{wang2024mechanical} applied DeePC to a planar soft robot, demonstrating safe and optimal control directly from input–output data without requiring explicit system identification. In subsequent work, Wang et al. \cite{wang2025velocity} proposed an incremental data-driven formulation that achieves robust and optimal control under unknown payload variations, mitigating performance degradation without relying on weighted datasets or disturbance estimators. Extending DeePC to spatial manipulation, Ouyang et al. \cite{ouyang2025direct} applied the framework to a three-dimensional (3D) cable-driven soft robotic arm, achieving both fixed-point regulation and trajectory tracking in 3D space. 
Despite these advances, DeePC operates in a receding-horizon manner, requiring the solution of a constrained optimization problem, typically a quadratic program, at every sampling instant \cite{coulson2019data}. For soft robotic systems that demand high sampling rates to capture fast transient behaviors and nonlinear deformation dynamics, this repeated re-optimization imposes a substantial computational burden, limiting real-time deployment on resource-constrained platforms.

To mitigate the computational burden in receding-horizon control frameworks, event-triggered control (ETC) mechanisms have been widely explored in optimization-based approaches, including MPC \cite{dang2023event} and Koopman-based predictive control~\cite{manaa2025koopman}. Recently, this paradigm has been extended to data-driven frameworks. For example, Wang et al. \cite{wang2025aperiodic} proposed an aperiodic data-driven predictive control scheme with terminal ingredients that guarantees recursive feasibility and stability. Similarly, Yang et al. \cite{yang2024event} developed an event-triggered strategy for multirate systems, where optimization updates are activated based on an accumulated error threshold.
However, existing event-triggered methods typically rely on static, heuristic triggering rules, such as fixed error thresholds or manually tuned performance factors \cite{wang2025aperiodic, yang2024event}. For soft robotic arms characterized by strongly nonlinear and time-varying dynamics, determining a globally optimal static threshold is inherently challenging. A conservative threshold often yields limited computational savings, whereas a relaxed threshold may lead to degraded tracking performance. Moreover, event-triggered DeePC remains relatively underexplored, particularly in scenarios that require adaptive and state-dependent update strategies to balance control accuracy and computational efficiency.

To address these challenges, we propose a \textbf{r}einforcement-\textbf{l}earning-based \textbf{e}vent-\textbf{t}riggered DeePC (RL-ET-DeePC) framework for soft robotic control. The proposed method integrates singular value decomposition (SVD)-based dimensionality reduction with a learned state-dependent triggering mechanism, where an RL agent determines whether to invoke the computationally expensive DeePC optimization or reuse previously computed control inputs (Fig.~\ref{fig:framework}). By adaptively balancing tracking accuracy and computational cost, the framework transforms conventional periodic DeePC into an event-driven predictive controller. The approach is validated through extensive simulations and hardware experiments on a 3D cable-driven soft robotic arm. Results demonstrate substantial reductions in optimization frequency compared to periodic DeePC, while achieving superior performance over static threshold-based event-triggered baselines. To promote reproducibility, the complete hardware design, source code, and simulation environment are publicly available online\footnote{Codes and designs: \url{https://github.com/chengoy30/RL-ET-DeePC-for-soft-robotic-arm-control.git}}.

\section{Hardware Design} \label{sec:hardware}
\begin{figure}[htbp]
    \centering
    \begin{subfigure}{0.15\textwidth}
        \centering
        \includegraphics[width=\linewidth]{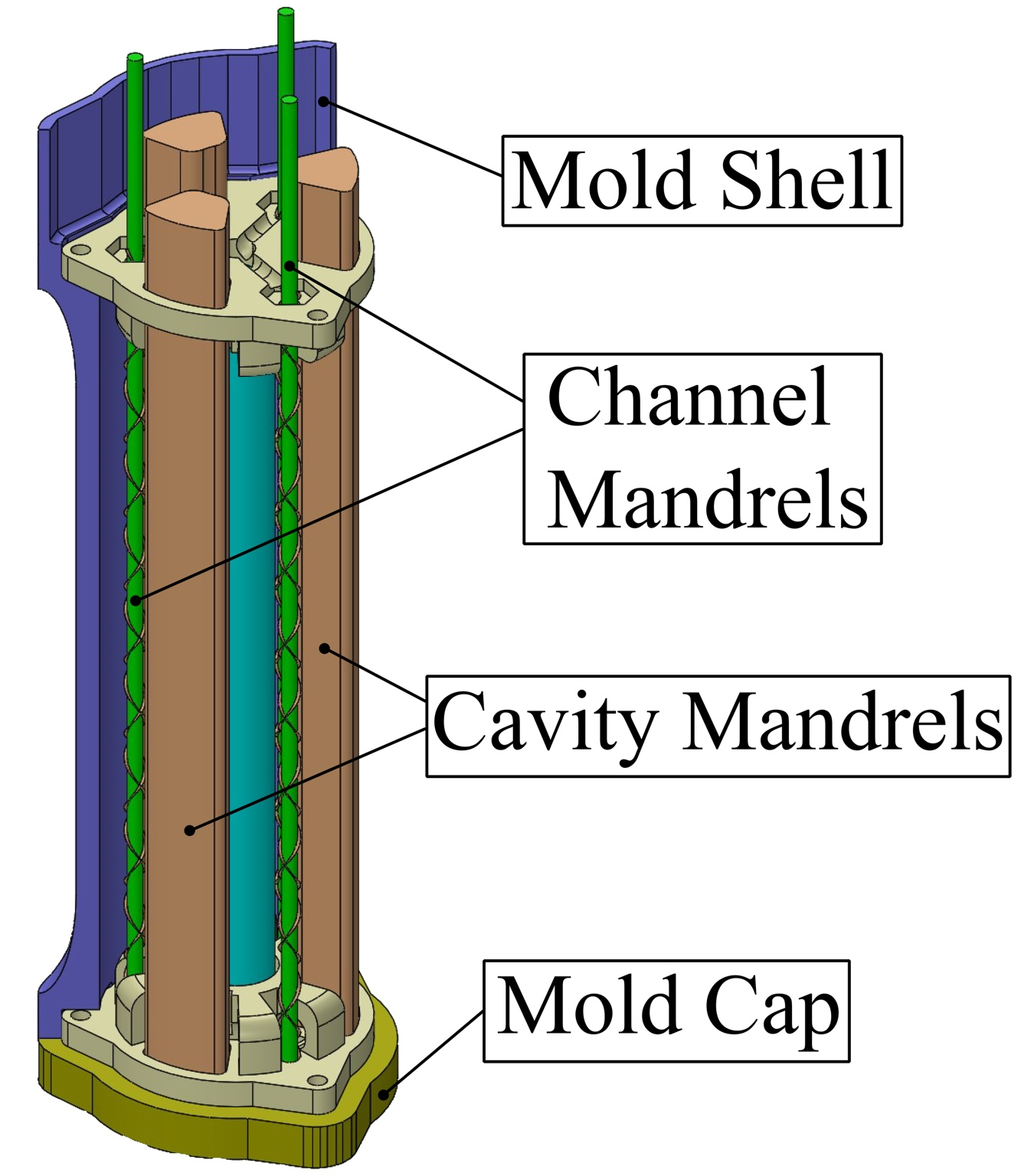}
        \caption{}
        \label{fig:softrobot_a}
    \end{subfigure}
    \begin{subfigure}{0.15\textwidth}
        \centering
        \includegraphics[width=\linewidth]{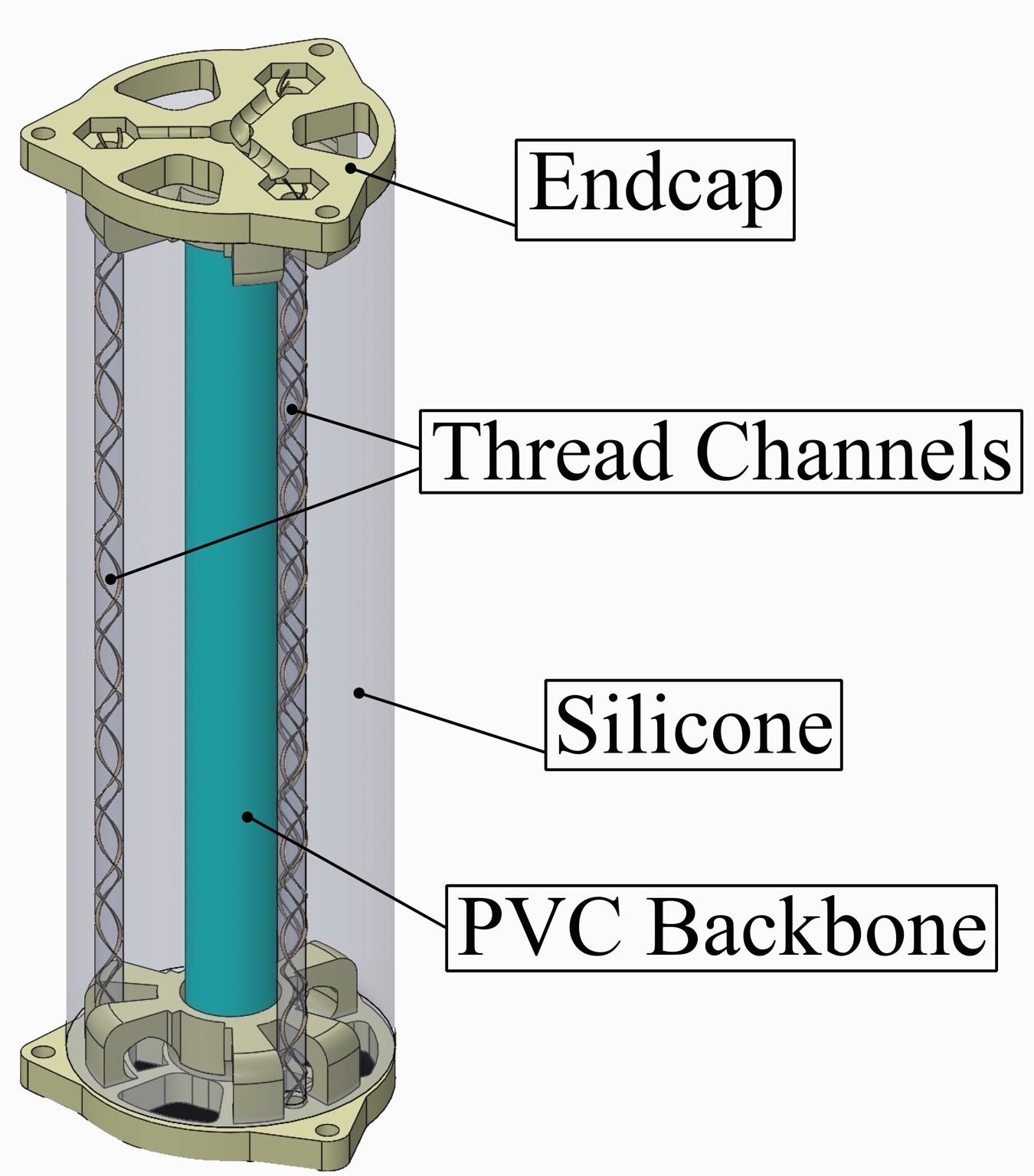}
        \caption{}
        \label{fig:softrobot_b}
    \end{subfigure}
    \begin{subfigure}{0.15\textwidth}
        \centering
        \includegraphics[width=\linewidth]{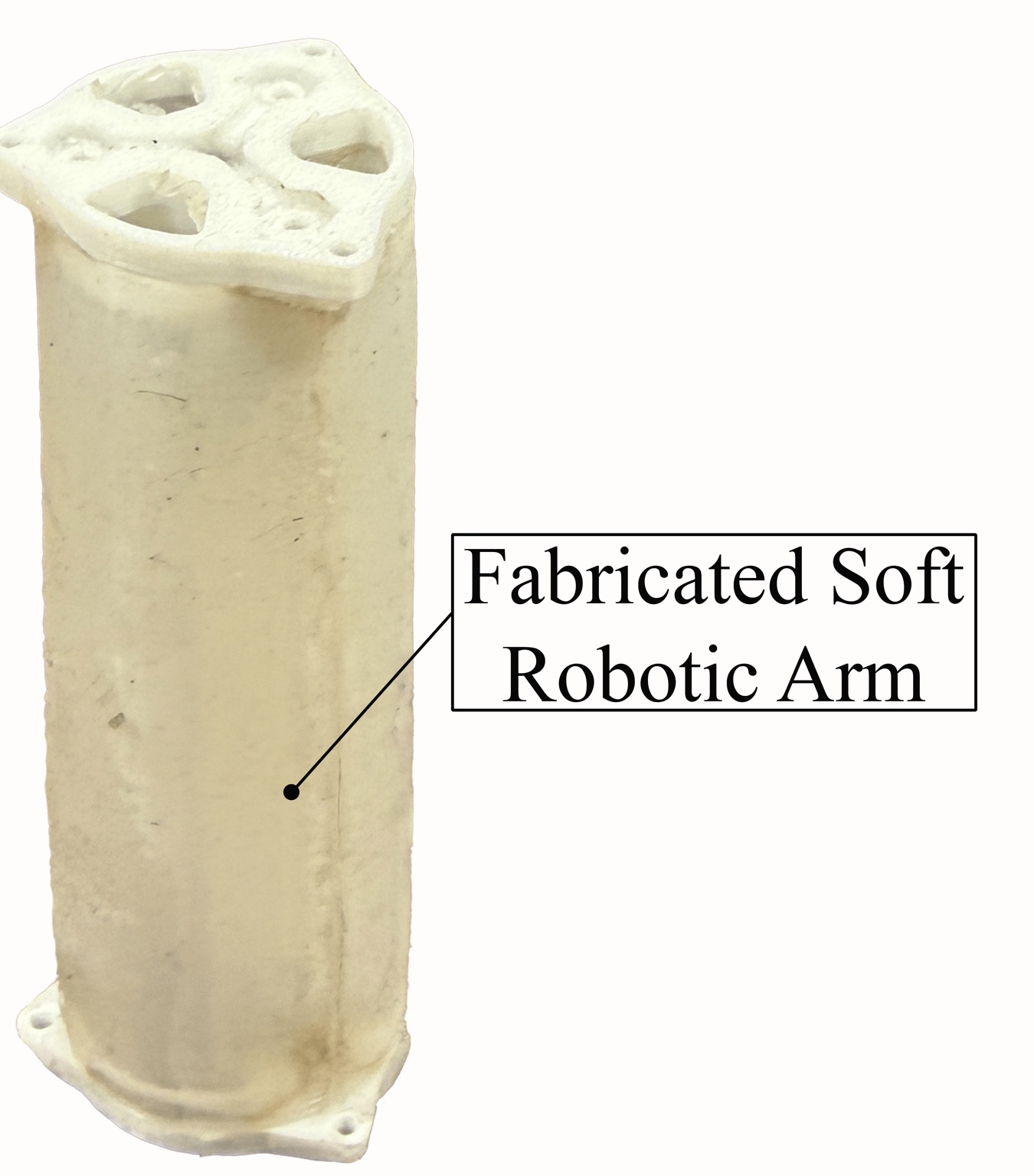} 
        \caption{}
        \label{fig:softrobot_c}
    \end{subfigure}
    \caption{3D cable-driven soft robotic arm: (a) casting molds and mandrels; (b) fabricated silicone module with embedded helical tendon-routing channels; (c) fully fabricated soft robotic arm.}
    \vspace{-10pt}
    \label{fig:softrobot_design}
\end{figure}

To evaluate the proposed RL-ET-DeePC framework, a 3D cable-driven soft robotic arm is employed, as depicted in Fig.~\ref{fig:softrobot_design}. The mandrels utilized during the casting process, the fabricated silicone module, and the completed prototype are shown in Fig.~\ref{fig:softrobot_a}, Fig.~\ref{fig:softrobot_b}, and Fig.~\ref{fig:softrobot_c}, respectively.  
The platform consists of a single soft continuum module (29\,mm diameter, 93\,mm length) constructed from 3D-printed rigid end caps, a flexible PVC backbone, and a cast silicone elastomer body (Ecoflex 00-50). Three Kevlar tendons are routed through internal helical channels embedded in the silicone body and anchored to the distal end cap. Their proximal ends exit the base and are actuated by stepper motors through a pulley transmission system. Differential cable actuation induces bending deformation, enabling continuous curvature motion in three-dimensional space.
The soft continuum structure exhibits pronounced nonlinear, hysteretic, and configuration-dependent dynamics due to material elasticity and tendon–structure coupling. These characteristics make accurate physics-based modeling difficult and computationally intensive for real-time control, thereby providing a representative benchmark for evaluating data-driven predictive control strategies.
 
From a control perspective, we formulate the cable-driven soft manipulator as a $3 \times 3$ multi-input multi-output plant. The input vector $u_k \in \mathbb{R}^3$ denotes motor step inputs that regulate tendon displacements, while the output vector $y_k \in \mathbb{R}^3$ denotes the Cartesian position $(x,y,z)$ of the end-effector. The reachable workspace spans approximately $\pm 44.4$\,mm in the $X$ and $Y$ directions and $-76.9$ to $-93.0$\,mm in the $Z$ direction.

\section{RL-based DeePC for Soft Robot Control} \label{sec:RL-based DeePC}

\subsection{DeePC Formulation for Soft Robot}
Following the standard DeePC formulation \cite{coulson2019data, wang2024mechanical}, we adopt the common assumption in data-driven control that the system trajectory can be locally approximated by a linear time-invariant (LTI) model around an operating point:
\begin{equation} \label{eq:LTI}
\begin{aligned}
x_{k+1} &= A x_k + B u_k \\
y_k &= C x_k + D u_k
\end{aligned}
\end{equation}
where $x_k \in \mathbb{R}^n$, $u_k \in \mathbb{R}^m$, and $y_k \in \mathbb{R}^p$ denote the state, input, and output vectors at time step $k$, respectively. Unlike model-based approaches that require explicit identification of matrices $(A \in \mathbb{R}^{n \times n}, B \in \mathbb{R}^{n \times m}, C \in \mathbb{R}^{p \times n}, D \in \mathbb{R}^{p \times m})$, DeePC directly leverages measured input–output trajectories to predict future system behavior, grounded in Willems’s Fundamental Lemma \cite{coulson2019data, willems2005note}.

To characterize the system behavior, we first collect an input-output trajectory of length $T$ offline. 
The collected dataset is denoted as 
$u^{\mathrm{d}} = \{u_1^{\mathrm{d}}, \dots, u_T^{\mathrm{d}}\}$ and 
$y^{\mathrm{d}} = \{y_1^{\mathrm{d}}, \dots, y_T^{\mathrm{d}}\}$. 
These data are arranged into block Hankel matrices. Specifically, the Hankel matrix $\mathcal{H}_{L}$ of depth $L$ is defined as follows:
\begin{equation}\label{eq:Hankel}
\mathcal{H}_{L} = 
\begin{bmatrix}
\mathcal{H}_{L}(u^\mathrm{d}) \\ \hline
\mathcal{H}_{L}(y^\mathrm{d})
\end{bmatrix}
=
\left[
  \begin{array}{cccc}
    u^\mathrm{d}_{1} & u^\mathrm{d}_{2} & \cdots & u^\mathrm{d}_{T-L+1} \\
    \vdots & \vdots & \ddots & \vdots \\
    u^\mathrm{d}_{L} & u^\mathrm{d}_{L+1} & \cdots & u^\mathrm{d}_{T} \\
    \hline
    y^\mathrm{d}_{1} & y^\mathrm{d}_{2} & \cdots & y^\mathrm{d}_{T-L+1} \\
    \vdots & \vdots & \ddots & \vdots \\
    y^\mathrm{d}_{L} & y^\mathrm{d}_{L+1} & \cdots & y^\mathrm{d}_{T}
  \end{array}
\right].
\end{equation}
According to Willems’s Fundamental Lemma~\cite{willems2005note}, if the input sequence $u^{\mathrm{d}}$ is persistently exciting of order $L+n$, then the column span of the Hankel matrix constructed from $(u^{\mathrm{d}}, y^{\mathrm{d}})$ characterizes all admissible trajectories of the underlying LTI system (Eq.~\eqref{eq:LTI}).
Based on this data-driven representation, any valid length-$L$ trajectory $(u_{[1, L]}, y_{[1, L]})$ can be expressed as a linear combination of the columns of the Hankel matrices using a decision vector $g \in \mathbb{R}^{T-L+1}$:
\begin{equation} \label{equ:trajectory}
\begin{bmatrix}
u_{[1,L]} \\
y_{[1,L]}
\end{bmatrix}
=
\begin{bmatrix}
\mathcal{H}_L \!\left(u^{\mathrm{d}}\right) \\
\mathcal{H}_L \!\left(y^{\mathrm{d}}\right)
\end{bmatrix}
g.
\end{equation}

However, the complexity of soft robots requires extensive offline data collection (i.e., large trajectory length $T$) to capture their dynamics. As a result, the Hankel matrix becomes high-dimensional, and the dimension of the decision variable $g$ grows accordingly, making real-time computation expensive. To address this issue, we apply singular value decomposition (SVD) to compress the data matrix \cite{zhang2023CSL}. Specifically, the concatenated input–output Hankel matrix (Eq.~\eqref{eq:Hankel}) is factorized into three components: left singular vectors, singular values, and right singular vectors, as follows: 
\begin{equation} \label{equ:svd}
\begin{bmatrix}
\mathcal{H}_L(u^{\mathrm{d}}) \\
\mathcal{H}_L(y^{\mathrm{d}})
\end{bmatrix}
=
\underbrace{\begin{bmatrix} W_1 & W_2 \end{bmatrix}}_{W\in \mathbb{R}^{q_{1}\times q_{1}}}
\underbrace{\begin{bmatrix} \Sigma_1 & 0 \\ 0 & \Sigma_2 \end{bmatrix}}_{\Sigma\in \mathbb{R}^{q_{1}\times q_{2}}}
\underbrace{\begin{bmatrix} V_1 & V_2 \end{bmatrix}}_{V^\top \in \mathbb{R}^{q_{2}\times q_{2}}},
\end{equation}
where $q_{1} = (m+p)L$ and $q_{2} = T-L+1$. $\Sigma$ contains singular values arranged in descending order.  $\Sigma_{1} \in \mathbb{R}^{r \times r}$ denote the diagonal matrix containing the top $r$ dominant singular values. The leading singular values capture the principal dynamical modes of the system.
Consequently, truncating to rank $r$ yields the reduced matrix
\begin{equation}
\bar{\mathcal{H}}_{L} = \mathcal{H}_{L} V_{1} = W_{1} \Sigma_{1},
\end{equation}
which preserves the dominant system behavior while significantly reducing dimensionality. Using this compressed representation, any valid trajectory can be approximated by
\begin{equation} \label{equ:H_bar}
\begin{bmatrix}
u_{[1,L]} \\
y_{[1,L]}
\end{bmatrix}
=
\bar{\mathcal{H}}_L
\bar{g},
\end{equation}
where $\bar{g} \in \mathbb{R}^r$ has a lower dimension compared to $g$.

\subsection{Regularized SVD-DeePC Optimization}
Due to measurement noise and modeling mismatch arising from the nonlinear and time-varying nature of soft robots, Eq.~\eqref{equ:H_bar} may not hold exactly \cite{wang2024mechanical}. To enhance feasibility and robustness, we formulate a regularized SVD-DeePC optimization problem. At each time step $t$, given the most recent input–output measurements $u_{\mathrm{ini}} = \{u_{t-T_{\mathrm{ini}}}, \dots, u_{t-1}\}$ and 
$y_{\mathrm{ini}} = \{y_{t-T_{\mathrm{ini}}}, \dots, y_{t-1}\}$, 
the controller solves the following optimization problem:
\begin{equation} \label{equ:deepc_opt}
\begin{aligned}
\min_{\bar{g},u,y,\sigma_y} \quad & 
\| y - y_r \|_Q^2 + \| u \|_R^2 
+ \lambda_y \| \sigma_y \|_2^2 
+ \lambda_g \| \bar{g} \|_2^2 \\
\text{subject to} \quad & 
\bar{\mathcal{H}}_{L} \bar{g} =
\begin{bmatrix}
u_{\mathrm{ini}} \\ u \\ y_{\mathrm{ini}} \\ y
\end{bmatrix}
+
\begin{bmatrix}
0 \\ 0 \\ \sigma_y \\ 0
\end{bmatrix}, \quad
u \in \mathcal{U}, \; y \in \mathcal{Y}.
\end{aligned}
\end{equation}
In Eq.~\eqref{equ:deepc_opt}, $y_{r}$ is the reference trajectory. $u = \{u_{t}, \dots, u_{t+N-1}\}$ and 
$y = \{y_{t}, \dots, y_{t+N-1}\}$ denote input and output prediction sequences over horizon $N$. $\mathcal{U}$ and $\mathcal{Y}$ are constraint sets.
The terms $\| \cdot \|_Q^2$ and $\| \cdot \|_R^2$ denote weighted quadratic costs on tracking error and control effort, respectively. The slack variable $\sigma_y$ relaxes the initial-condition constraint to account for measurement noise. The regularization terms $\lambda_g \| \bar{g} \|_2^2$ and $\lambda_y \| \sigma_y \|_2^2$ mitigate overfitting to noisy data and suppress excessively large decision variables, thereby improving numerical stability. Let the optimal predicted control sequence be 
$\mathbf{u}_f^* = \{u_t^*, \dots, u_{t+N-1}^*\}$. 
In conventional receding-horizon schemes, only the first control input $u_t^*$ is applied before re-solving the optimization problem at time $t+1$. In contrast, this work introduces an event-triggered mechanism that selectively reuses subsequent elements of $\mathbf{u}_f^*$ to defer expensive re-optimization steps, as described in the next subsections.

\subsection{RL-based Event-Triggered DeePC}
To alleviate the computational burden of solving the optimization problem (Eq.~\eqref{equ:deepc_opt}) at every time step, we propose an event-triggered framework governed by a deep reinforcement learning (DRL) agent. As illustrated in Fig.~\ref{fig:framework}, the proposed approach adopts an event-driven architecture in which the RL policy determines \textit{when} to invoke the computationally expensive DeePC optimization, while the soft robot receives a control input at every sampling instant, either freshly computed or retrieved from the predictive buffer.
At each time step $t$, the DRL agent $\pi_\theta$, parameterized by neural network weights $\theta$, selects a binary action $a_t \in \{0, 1\}$, determining whether to trigger a re-optimization. If $a_t = 1$, the regularized SVD-DeePC optimization problem in Eq.~\eqref{equ:deepc_opt} is solved to update the decision variable $\bar{g}$ and a new optimal control sequence. The first control input $u_t^*$ of the updated sequence is then applied to the robot. If $a_t = 0$, the optimization is skipped to reduce computational load, and the controller instead applies the corresponding input from the previously computed optimal sequence stored in a buffer $\mathcal{U}_{\mathrm{buf}}$. Specifically, let $t_{\mathrm{last}} \leq t$ denote the most recent time step at which the optimization was triggered, the control input applied at the current time $t$ is then $u_t=u_{t|t_{last}}^{\ast}$, which is the element from the previously computed sequence that corresponds to the current time step. 
This strategy effectively utilizes the remaining portion of the prediction horizon, transforming conventional receding-horizon control into an event-driven re-optimization mechanism with adaptive update frequency. 
The interaction between the agent and the environment proceeds in episodes of maximum length $T_{\max}$ time steps, after which the simulation environment is reset.

We formulate the event-triggering mechanism as a Markov Decision Process (MDP) defined by the tuple $(\mathcal{S}, \mathcal{A}, \mathcal{R}, \mathcal{P}, \gamma)$:
\begin{itemize}
    \item \textbf{State space:} The state $s_t \in \mathcal{S}$ captures the tracking performance and its temporal evolution. We define 
    $s_t = \begin{bmatrix}e_t^{\top}, \Delta e_t^{\top} \end{bmatrix}^{\top}$, where $e_t = r_t - y_t$ is the tracking error with $r_t$ and $y_t$ denoting the reference output and system output, $\Delta e_t = e_t - e_{t-1}$ represents its variation. This compact representation provides sufficient information for the agent to assess current tracking quality and short-term error trends.

    \item \textbf{Action space:} The action space $a_t \in \mathcal{A}$ is discrete and binary, $\mathcal{A} = \{0,1\}$. 
    The action $a_t = 1$ triggers the SVD-DeePC optimization update, while $a_t = 0$ applies the buffered control input without re-optimization.

    \item \textbf{Reward function:} The immediate reward balances tracking performance and computational usage:
    $R_t = - \left( \| e_t \|_W^2 + \rho \cdot a_t \right)$,
    where $\|e_t\|_W^2$ penalizes tracking error weighted by the positive-definite matrix $W$, and $a_t$ represents the computational cost of triggering. 
    The parameter $\rho > 0$ controls the trade-off between tracking accuracy and computational cost. A larger $\rho$ penalizes triggering more heavily, encouraging fewer optimization updates, whereas a smaller $\rho$ prioritizes tracking performance. Unlike fixed-threshold methods~\cite{liu2024two}, $\rho$ does not directly determine when to trigger; instead, it shapes the overall performance--computation balance while the RL agent learns a state-dependent triggering policy.
    
    \item \textbf{Transition probability:} The state transition is governed by the closed-loop robotic dynamics. Since the system dynamics are unknown, $\mathcal{P}$ is not explicitly modeled and is instead learned implicitly through interaction with the environment.

    \item \textbf{Discount factor:} The discount factor $\gamma \in (0,1)$ determines the relative importance of future rewards. 
    A larger $\gamma$ encourages long-term performance optimization, while a smaller $\gamma$ emphasizes immediate tracking and computational efficiency.
\end{itemize}

In this work, we emphasize the proposed event-triggered control framework rather than any specific RL algorithm. To demonstrate generality, we evaluate representative algorithms from value-based (DQN \cite{mnih2015human}), policy-gradient (PPO \cite{schulman2017proximal}), and actor-critic (A2C \cite{mnih2016asynchronous}) families under a discrete binary action setting. Importantly, the event-triggering mechanism itself is algorithm-agnostic and can be readily integrated with other RL approaches. 

\vspace{-5pt}
\subsection{Supervisory Override Mechanism} \label{sec:safety}
To enhance operational robustness, we introduce a supervisory override layer (Fig.~\ref{fig:framework}) that complements the RL agent’s decision. This mechanism ensures the availability of a valid control sequence and prevents excessive open-loop execution. Specifically, if the control buffer is empty ($\mathcal{U}_{\mathrm{buf}} = \emptyset$), the optimization is triggered unconditionally so that a feasible control plan is always maintained. In addition, if the elapsed time since the last update satisfies $t - t_{\mathrm{last}} \geq N - 1$, a forced trigger is issued, since the buffer can store at most $N - 1$ remaining control inputs. Accordingly, the effective triggering signal $\delta_t \in \{0,1\}$ is defined as:
\begin{equation}\label{eq.supervisor}
\delta_t =
\begin{cases}
1, & \text{if } 
a_t = 1 
\;\lor\;
\mathcal{U}_{\mathrm{buf}} = \emptyset
\;\lor\;
t - t_{\mathrm{last}} \geq N-1, \\
0, & \text{otherwise}.
\end{cases}
\end{equation}

A re-optimization is executed whenever $\delta_t = 1$. This supervisory logic bounds the maximum open-loop interval by the prediction horizon $N$ and guarantees periodic controller refresh independent of the learned policy. Note that this override layer acts as a supervisory safeguard within the control architecture rather than a formal stability certificate. 
%
The details of the proposed RL-ET-DeePC framework are summarized in Algorithm~\ref{alg:rl_deepc}.


        


\begin{figure}[!ht]
\vspace{-10pt}
\removelatexerror
\scalebox{0.9}{
\begin{algorithm*}[H]
\SetAlFnt{\small}
\SetKwInOut{Parameter}{Parameters}
\SetKwInOut{Output}{Output}
\caption{RL-ET-DeePC Framework}
\label{alg:rl_deepc}
\SetAlgoLined

\Parameter{$\bar{\mathcal{H}}_{L}$, $\pi_{\theta}$, $N$, $T_{\max}$}

Initialize control buffer $\mathcal{U}_{\mathrm{buf}} \gets \emptyset$, $t_{\mathrm{last}} \gets 0$ \\

\For{each training episode}{
Reset environment and obtain initial state $s_0$ \\

\For{$t = 0$ to $T_{\max}-1$}{
Select action $a_t \sim \pi_{\theta}(\cdot | s_t)$ \\


Compute $\delta_t$ using Eq.~\eqref{eq.supervisor} \\

\eIf{$\delta_t = 1$}{
Solve SVD-DeePC optimization (Eq.~\eqref{equ:deepc_opt}) \\
Obtain optimal sequence $\mathbf{u}_f^{\ast} = \{u_t^{\ast}, \dots, u_{t+N-1}^{\ast}\}$ \\
Apply $u_t \gets u_t^{\ast}$ \\
Update buffer $\mathcal{U}_{\mathrm{buf}} \gets \{u_{t+1}^{\ast}, \dots, u_{t+N-1}^{\ast}\}$ \\
Update $t_{\mathrm{last}} \gets t$
}{
Apply next control input from $\mathcal{U}_{\mathrm{buf}}$
}

Observe next state $s_{t+1}$ and reward $R_t$ \\
Store transition $(s_t, a_t, R_t, s_{t+1})$ and update $\pi_{\theta}$ \\
$s_t \gets s_{t+1}$
}
}
\end{algorithm*}}
\vspace{-10pt}
\end{figure}

\vspace{-5pt}
\section{Simulation Results and Analysis} \label{sec:simulation}
\subsection{Simulation Setup}
To evaluate the proposed RL-ET-DeePC framework, we developed a high-fidelity simulation environment based on the nonlinear kinematic model of a modular cable-driven soft robotic arm presented in \cite{qi2024design}. The simulator is implemented with a Gym-compatible interface \cite{brockman2016openai}, enabling seamless integration with standard RL algorithms.
All RL agents share the same state representation, reward formulation, and training environment to ensure a fair comparison. Specifically, the weighting matrix in the reward function is set to $W = 3.33 I$. For network architecture, all methods employ a single hidden layer with 128 neurons and a discount factor $\gamma = 0.98$.
For the policy-gradient methods (A2C and PPO), the actor and critic learning rates are set to $10^{-3}$ and $10^{-2}$, respectively, with a Generalized Advantage Estimation (GAE) parameter $\lambda = 0.95$ \cite{schulman2017proximal}. A2C uses an entropy coefficient of $0.01$ to promote exploration, while PPO adopts a clipping range $\epsilon = 0.2$ and performs 10 update epochs per iteration \cite{mnih2016asynchronous}. For the value-based DQN algorithm, the learning rate is set to $2 \times 10^{-3}$ with an exploration rate $\epsilon = 0.02$. The target network is updated every 10 steps. The replay buffer capacity is 5000 transitions (minimum size 500) with a batch size of 64 \cite{mnih2015human}. All RL agents are trained for 50,000 time steps, organized into episodes of 200 steps each.

To construct the non-parametric data model required for DeePC, we performed an initial data-collection phase in the kinematic simulation environment \cite{qi2024design} using a ``ramp-and-hold'' excitation signal. A dataset comprising 1200 input-output samples was recorded to construct the Hankel matrices, utilizing a trajectory length of $N=25$, an initialization horizon of $T_{\mathrm{ini}}=20$, and an SVD truncation rank of $r = 600$. Although the underlying model is kinematic, the data-driven formulation of DeePC treats the system response as a black-box input-output map, effectively learning the local linearized behavior along the trajectory \cite{coulson2019data}.

The underlying predictive controller was configured to prioritize high-precision tracking while maintaining numerical stability. The weighting matrices in the optimization objective were set to $Q = 800 I$ and $R = 10^{-5} I$, respectively. To handle measurement noise and ensure the regularity of the operator $\bar{g}$, the regularization parameters were chosen as $\lambda_g = 300$ and $\lambda_y = 1500$. Furthermore, physical constraints were strictly imposed on the system: the control inputs (stepper motor steps) were bounded within $[-1200, 1200]$ steps across all channels. Similarly, the system state constraints were set to $[-100, 100]$ mm for the first two dimensions, while the third dimension was constrained to $[-100, 0]$ mm to reflect the physical workspace limits of the soft arm.

\vspace{-5pt}
\subsection{Accuracy-Triggering Trade-off Analysis}
To evaluate the proposed RL-ET-DeePC framework, each trained policy is tested on a circular reference trajectory identical to the one used during training, with each test episode lasting 200 time steps. Table~\ref{tab:trigger_comparison} quantitatively compares RL-ET-DeePC (DQN, A2C, PPO) with periodic DeePC. The results clearly demonstrate that the proposed event-triggered mechanism enables substantial reductions in triggering frequency and total decision time (defined as the cumulative time spent by the DeePC controller in solving the optimization problems throughout the entire test duration) while maintaining comparable tracking accuracy. To isolate the effect of the RL algorithm and the penalty parameter $\rho$ from training stochasticity, each RL configuration is trained and evaluated over 5 independent random seeds, and the reported values are given as mean ± standard deviation. The consistently low variance across seeds confirms that the observed differences in RMSE and trigger rate stem from the choice of algorithm and $\rho$ rather than from training randomness.
For the periodic DeePC controller, the trigger rate remains at 100\%, yielding an RMSE of 0.144 mm and a total decision time of 8.94 s, corresponding to an average of approximately 0.045 s per optimization call. In contrast, all RL-based variants significantly reduce the trigger rate as the penalty parameter $\rho$ increases. For example, PPO reduces the trigger rate from 100\% to 44.4\% at $\rho = 0.5$ and further to 33.9\% at $\rho = 1.0$, corresponding to a 66.1\% reduction in optimization calls. Despite this reduction, the tracking accuracy remains within an acceptable range, with RMSE values of 0.230 mm and 0.288 mm, respectively. A similar trend is observed for DQN and A2C, confirming that the learned policy effectively balances accuracy and triggering frequency.
Among all methods, PPO consistently achieves the most favorable trade-off between control performance and triggering efficiency. At $\rho = 0.5$, PPO reduces the decision time to 3.95 s (approximately 56\% lower than periodic DeePC) while maintaining moderate tracking accuracy. This indicates that PPO is better able to learn when re-optimization is necessary, avoiding redundant updates while preserving closed-loop stability. In the following sections, PPO is therefore adopted as the representative RL algorithm for further analysis and hardware validation.

\begin{table}[t]
\centering
\caption{Performance comparison of RL-ET-DeePC (DQN, A2C, PPO) and periodic DeePC. Values are reported as mean ± standard deviation over 5 random seeds.}
\label{tab:trigger_comparison}

\footnotesize
\setlength{\tabcolsep}{2.8pt}
\renewcommand{\arraystretch}{1.15}

\begin{tabular}{@{}lccccc@{}}
\toprule
\textbf{Method}
& $\boldsymbol{\rho}$
& \textbf{Return}
& \textbf{Trigger (\%)}
& \textbf{RMSE (mm)}
& \textbf{Time (s)} \\
\midrule

\multirow{3}{*}{DQN}
& 0.1 & $-118.19 \mathbin{\pm} 0.98$ & $82.8 \mathbin{\pm} 1.5$ & $0.157 \mathbin{\pm} 0.005$ & $7.33 \mathbin{\pm} 0.17$ \\
& 0.5 & $-171.60 \mathbin{\pm} 1.70$ & $45.1 \mathbin{\pm} 1.3$ & $0.242 \mathbin{\pm} 0.003$ & $4.01 \mathbin{\pm} 0.14$ \\
& 1.0 & $-213.00 \mathbin{\pm} 1.50$ & $39.2 \mathbin{\pm} 1.2$ & $0.266 \mathbin{\pm} 0.006$ & $3.49 \mathbin{\pm} 0.10$ \\
\midrule

\multirow{3}{*}{A2C}
& 0.1 & $-118.43 \mathbin{\pm} 1.09$ & $99.0 \mathbin{\pm} 0.4$ & $0.144 \mathbin{\pm} 0.002$ & $8.75 \mathbin{\pm} 0.05$ \\
& 0.5 & $-190.09 \mathbin{\pm} 1.91$ & $88.1 \mathbin{\pm} 2.0$ & $0.161 \mathbin{\pm} 0.006$ & $7.79 \mathbin{\pm} 0.16$ \\
& 1.0 & $-244.84 \mathbin{\pm} 6.97$ & $59.6 \mathbin{\pm} 3.4$ & $0.238 \mathbin{\pm} 0.016$ & $5.29 \mathbin{\pm} 0.30$ \\
\midrule

\multirow{3}{*}{PPO}
& 0.1 & $-129.65 \mathbin{\pm} 1.01$ & $100.0 \mathbin{\pm} 0.0$ & $0.144 \mathbin{\pm} 0.001$ & $8.89 \mathbin{\pm} 0.07$ \\
& 0.5 & $-162.97 \mathbin{\pm} 0.90$ & $44.4 \mathbin{\pm} 0.5$ & $0.230 \mathbin{\pm} 0.002$ & $3.95 \mathbin{\pm} 0.04$ \\
& 1.0 & $-207.23 \mathbin{\pm} 2.74$ & $33.9 \mathbin{\pm} 0.5$ & $0.288 \mathbin{\pm} 0.005$ & $3.02 \mathbin{\pm} 0.07$ \\
\midrule

DeePC
& --
& --
& $100.0$
& $0.144 \mathbin{\pm} 0.001$
& $8.94 \mathbin{\pm} 0.07$ \\
\bottomrule
\end{tabular}
\vspace{-14pt}
\end{table}

\begin{figure}[!ht]
  \centering
  \includegraphics[width=0.85\columnwidth]{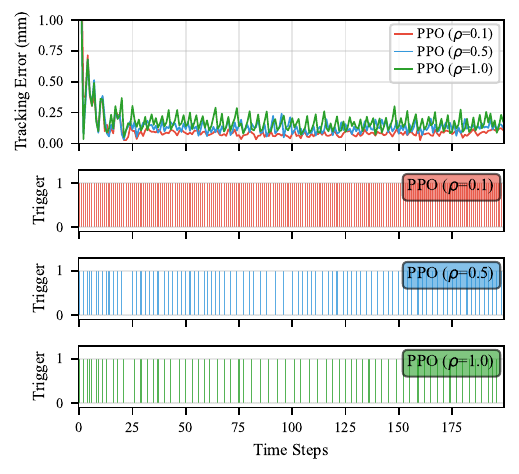}
  \vspace{-10pt}
  \caption{Tracking error (RMSE) and trigger actions of PPO-based RL-ET-DeePC under different $\rho$ values.}
  \label{fig:rho_effect}
  \vspace{-10pt}
\end{figure}

Fig.~\ref{fig:rho_effect} further illustrates the adaptive triggering behavior of PPO under different $\rho$ values. The top subplot shows the tracking error trajectories, where all three settings converge rapidly and remain bounded. As $\rho$ increases, the average tracking error slightly increases, reflecting the stronger penalty on triggering. The lower subplots display the corresponding trigger instants. For $\rho = 0.1$, triggering is dense throughout the horizon, closely resembling periodic control. In contrast, for $\rho = 0.5$ and $\rho = 1.0$, triggering is more concentrated during the early transient phase, when tracking errors are larger, and becomes significantly sparser once the system enters steady state. This behavior indicates that the learned policy prioritizes re-optimization during periods of rapid state variation and suppresses redundant updates when the system operates near the desired trajectory.
Overall, the results validate that RL-ET-DeePC learns an adaptive event-triggering strategy that reduces computational burden while preserving tracking performance. The parameter $\rho$ provides an intuitive mechanism to tune the accuracy–efficiency trade-off, enabling flexible deployment depending on real-time computational constraints.

\vspace{-5pt}
\subsection{Generalization to Unseen Trajectories}
To further evaluate the robustness and generalization capability of the proposed RL-ET-DeePC framework, we test the trained policies on an unseen multi-step reference trajectory that differs from those encountered during training. Although the DeePC controller operates on the Cartesian tip position (x,y,z) as defined in Section~\ref{sec:hardware}, the results are presented in terms of bending angle and bending direction. This trajectory contains multiple abrupt changes in both bending angle and bending direction, thereby challenging the controller under rapid transient variations.

\begin{figure}[!ht]
  \centering
  \includegraphics[width=0.85\columnwidth]{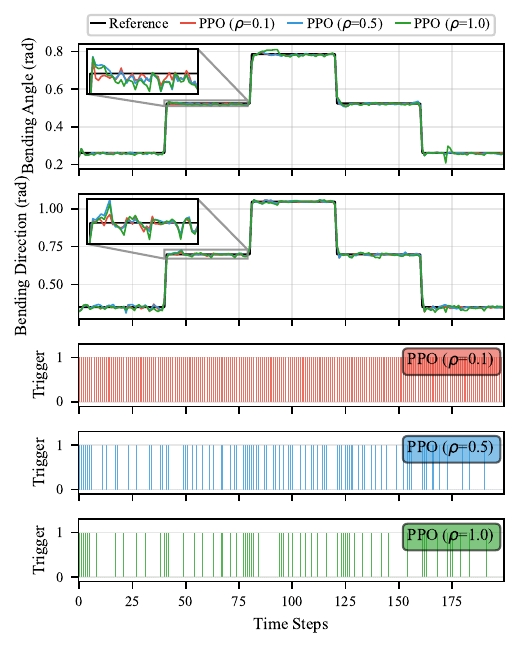}
   \vspace{-10pt}
\caption{Tracking performance and event-triggering actions of PPO-based RL-ET-DeePC on an unseen multi-step trajectory.}
  \label{fig:unseen}
  \vspace{-14pt}
\end{figure}

Fig.~\ref{fig:unseen} shows the tracking performance and corresponding triggering behavior of PPO-based RL-ET-DeePC under different $\rho$ values. The first two subplots show the bending angle and bending direction responses. All configurations successfully track the multi-step reference with small steady-state error and limited overshoot, demonstrating that the learned triggering strategy generalizes beyond the training distribution.
During sharp reference transitions (e.g., around time steps 40, 80, and 160), temporary tracking deviations are observed, particularly for larger $\rho$. This behavior is consistent with the stronger penalty imposed on triggering. The lower three subplots display the trigger instants. For $\rho = 0.1$, triggering remains dense and closely resembles periodic control. In contrast, for $\rho = 0.5$ and $\rho = 1.0$, triggering events become more concentrated around reference switching points, while remaining sparse during steady-state segments. This confirms that the learned policy activates re-optimization primarily when significant state deviations occur, and suppresses redundant updates during steady operation. 
Overall, these results demonstrate that RL-ET-DeePC not only achieves an effective accuracy–efficiency trade-off in nominal conditions, but also maintains reliable performance when exposed to previously unseen trajectories.

\vspace{-5pt}
\section{Hardware Experiments on the Soft Robotic Arm} \label{sec:experiment}
\subsection{Experimental Platform}
\begin{figure}[!ht]
    \centering
    \includegraphics[width=0.35\textwidth]{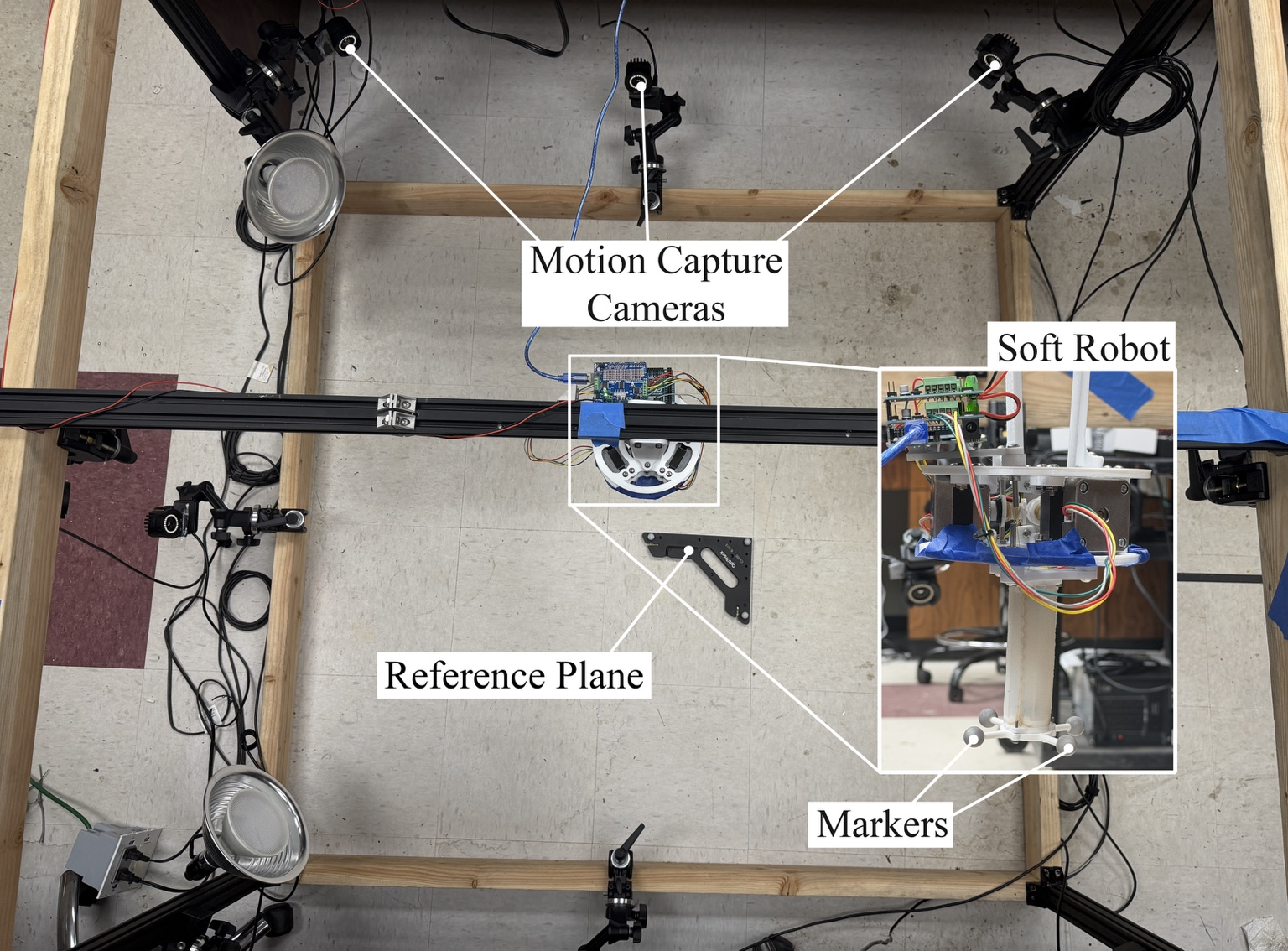}
    \caption{Experimental setup with motion capture cameras and a test frame.}
    \label{fig:hardware_setup}
    \vspace{-14pt}
\end{figure}

Real-world validation was conducted on the custom-built cable-driven soft robotic arm (Sec.~\ref{sec:hardware}) shown in Fig.~\ref{fig:hardware_setup}. The system consists of a single soft continuum module actuated by stepper motors through an Arduino Mega 2560 and Adafruit Motor Shield V2 drivers. The actuation operates in open-loop without encoder feedback. 
To provide accurate state measurement, the end-effector position is captured in real time using an OptiTrack motion capture system equipped with 12 FLEX13 cameras. The RL-ET-DeePC controller runs on a host PC, which performs the SVD-DeePC optimization and PPO inference, and transmits motor commands to the microcontroller via serial communication. 

\vspace{-5pt}
\subsection{Statistical Performance Evaluation}
\begin{figure}[!ht]
  \centering
  \includegraphics[width=0.90\columnwidth]{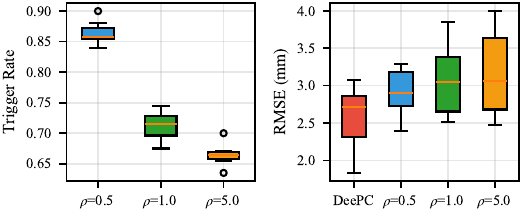}
  \caption{Tracking RMSE and trigger rate under different $\rho$.}
  \label{fig:boxplot}
  \vspace{-14pt}
\end{figure}

To assess the control performance in experiments, we compare periodic DeePC with PPO-based RL-ET-DeePC under three penalty coefficients: $\rho \in \{ 0.5, 1.0, 5.0\}$. Each configuration is tested over 10 independent trials. Fig.~\ref{fig:boxplot} presents box-and-whisker plots of the RMSE and the trigger rate. The periodic DeePC serves as a high-accuracy baseline with a median RMSE of approximately 2.75 mm and a 100\% trigger rate. 
When $\rho = 0.5$, RL-ET-DeePC maintains comparable tracking accuracy while reducing the trigger rate to approximately 86\%. Increasing the penalty to $\rho = 5.0$ further decreases the median trigger rate to about 66\%, achieving a 34\% reduction in optimization frequency relative to periodic DeePC. 
Although larger $\rho$ slightly increases RMSE variance, the overall tracking performance remains stable across all trials. These results confirm that the learned RL policy successfully transfers from simulation to hardware and preserves the desired accuracy–efficiency trade-off.
Note that the effective range of $\rho$ differs between simulation and hardware experiments due to modeling mismatch and disturbances; therefore, the penalty values are tuned separately for each setting to ensure comparable operating conditions.

\vspace{-5pt}
\subsection{Representative Trajectory Tracking}
\begin{figure}[!t]
    \centering
    \includegraphics[width=0.85\columnwidth]{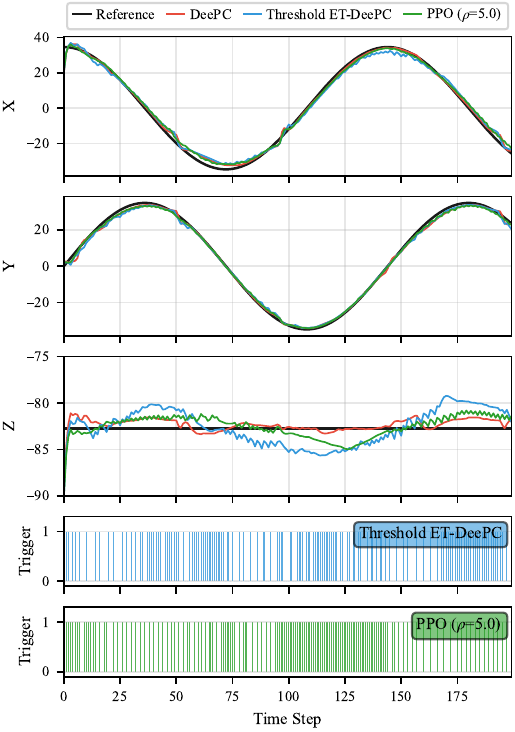}
    \vspace{-10pt}
    \caption{Real-robot tracking performance along the $X$, $Y$, and $Z$ axes and the corresponding triggering signals for periodic DeePC, threshold-based ET-DeePC, and PPO-based RL-ET-DeePC ($\rho=5.0$).}
    \label{fig:compare_y_target}
    \vspace{-14pt}
\end{figure}

To illustrate the closed-loop behavior, Fig.~\ref{fig:compare_y_target} shows the time-domain spatial path and the time-domain coordinate tracking responses along the $X$, $Y$, and $Z$ axes for a representative trial. A threshold-based event-triggered DeePC method, which triggers re-optimization whenever the tracking error exceeds a predefined threshold, is included as a baseline for comparison. To ensure a fair comparison, the error threshold of this baseline was tuned so that its average trigger rate closely matches that of the PPO-based RL-ET-DeePC at $\rho = 5.0$; the two methods therefore operate under a comparable computational budget, and any difference in tracking accuracy reflects where re-optimization is allocated rather than how often it occurs.
Both periodic DeePC and RL-ET-DeePC accurately track the sinusoidal reference in the $X$ and $Y$ directions with minimal phase lag. The $Z$-axis exhibits small oscillations due to the intrinsic compliance and open-loop vertical holding of the cable-driven structure; however, the spatial trajectory remains well regulated.
Comparing the triggering behavior reveals a key difference. The threshold-based method generates bursts of updates whenever the tracking error exceeds the preset bound, resulting in irregular triggering patterns. In contrast, the PPO-based controller produces structured and state-dependent triggering: updates are concentrated during dynamic transitions and significantly reduced during steady-state motion.
This behavior indicates that our proposed  RL-ET-DeePC has learned a context-aware update strategy rather than relying on a fixed error magnitude. Even under aggressive penalization ($\rho = 5.0$), the learned policy maintains closed-loop stability without noticeable trajectory degradation.

\begin{figure*}[!ht]
    \centering
    \includegraphics[width=0.75\textwidth]{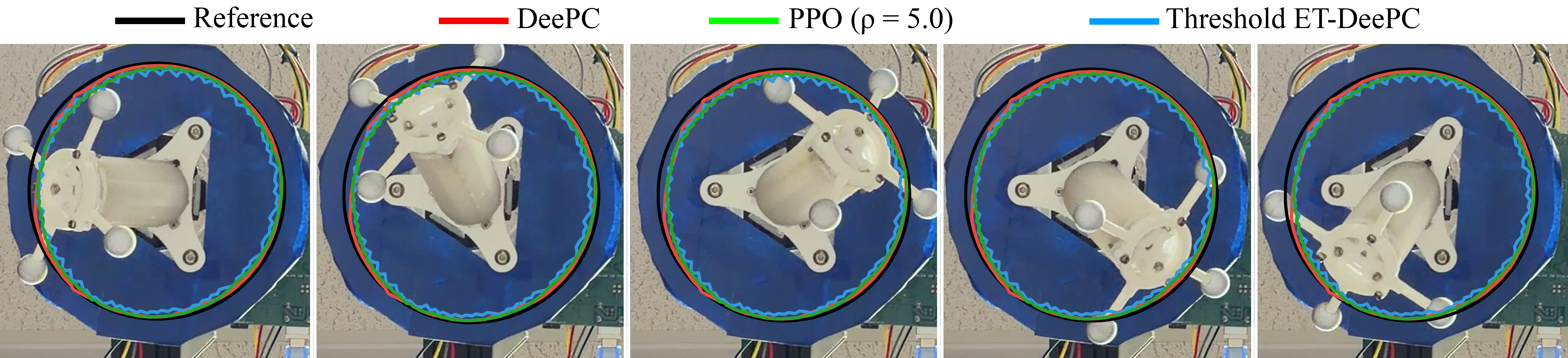}
\caption{Tracking performance on the physical arm. The black curve is the reference circular trajectory. Compared with periodic DeePC and PPO-based RL-ET-DeePC ($\rho=5.0$), the threshold-based ET-DeePC exhibits visibly larger tracking deviations.}
    \label{fig:real_3d_trajectory}
    \vspace{-14pt}
\end{figure*}

Fig.~\ref{fig:real_3d_trajectory} shows representative overhead snapshots of spatial tracking on the physical platform, where the measured end-effector trajectories are overlaid on the reference circular path. 
From the spatial comparison, both periodic DeePC and RL-ET-DeePC closely follow the reference circle with only minor deviations, indicating that reducing optimization frequency does not noticeably degrade the steady-state accuracy. In contrast, the threshold-based method exhibits visibly larger radial deviations, particularly in high-curvature regions of the trajectory. This suggests that fixed error-bound triggering may delay necessary updates during dynamic transitions.
Importantly, the PPO-based policy achieves tracking performance nearly indistinguishable from periodic DeePC while significantly reducing optimization calls. The smoother and more uniformly distributed deviations further indicate that the learned triggering strategy adapts to motion context rather than reacting solely to instantaneous error magnitude. These results confirm that RL-ET-DeePC preserves geometric tracking fidelity on hardware while improving computational efficiency.

\vspace{-5pt}
\section{Conclusion \& Discussion}
In this paper, we proposed an RL-ET-DeePC framework for computationally efficient and high-precision control of cable-driven soft robotic arms. By integrating SVD-based DeePC with a DRL agent to adaptively govern optimization updates, the approach alleviates the computational burden of periodic DeePC while preserving tracking accuracy. Extensive simulation and hardware experiments demonstrated that the learned policy achieves a more consistent and adaptive accuracy–cost trade-off than static threshold-based event-triggering. We emphasize that the contribution lies in formulating event-triggering as a learned, state-dependent decision and the framework is agnostic to the specific RL algorithms.

Although all experiments use a single-module, three-tendon arm, the framework is largely platform-agnostic: both the DeePC representation and the RL policy rely only on input-output data and a scalar reward, without assuming a specific actuation mechanism or morphology. It therefore extends in principle to other soft robots, such as pneumatically actuated continuum arms and multi-segment systems, provided a persistently exciting dataset and a low-dimensional I/O interface that keep the Hankel matrices and SVD rank $r$ tractable. Morphologies with higher-dimensional or more strongly coupled dynamics would enlarge the data requirement and rank $r$, while stronger hysteresis would complicate excitation design and widen the sim-to-real gap.

This gap is visible in our results: the optimization reduction drops from 66\% in simulation to 34\% on hardware. The simplified kinematic simulator omits actuation delay, friction, and cable hysteresis, and the open-loop, encoder-free platform injects state variations that the policy reads as conditions warranting re-optimization, raising the effective trigger rate. Consequently the effective range of $\rho$ shifts between domains and must be retuned on hardware. This burden may be reduced by normalizing the reward cost term so that $\rho$ becomes more domain-invariant, by domain randomization over unmodeled dynamics, and by online data updating. Future directions include dynamic-aware simulation, online data updating, and tube-based DeePC formulations for stronger stability guarantees.

\vspace{-10pt}
\bibliography{ref}
\bibliographystyle{IEEEtran}

\end{document}